\documentclass[runningheads]{final_template/svproc}
\usepackage{graphicx}
\usepackage{color}
\usepackage{amsmath}
\usepackage{caption}
\usepackage{subcaption}
\usepackage{wrapfig}
\begin{document}
\newcommand{\braces}[1]{\left\{ #1 \right\}}
\newcommand{\leftbrace}[1]{\left\{ #1 \right.}
\newcommand{\brackets}[1]{\left[ #1 \right]}
\newcommand{\paren}[1]{\left( #1 \right)}
\newcommand{\evalat}[1]{\left. #1 \right|}
\newcommand{\abs}[1]{\left| #1 \right|}
\newcommand{\norm}[1]{\left\|#1\right\|}
\newcommand{\mat}[1]{\ensuremath{\left[#1\right]}}
\newcommand{\bs}[1]{\ensuremath{\boldsymbol{#1}}}
\newcommand{\matb}[1]{\ensuremath{\mat{\bs{#1}}}}

% large block symbols
\newcommand{\RNum}[1]{\textbf{\uppercase\expandafter{\romannumeral #1\relax}}}
\newcommand{\one}{\RNum{1}}
\newcommand{\two}{\RNum{2}}
\newcommand{\bfJ}{\mathbf{J}}
\newcommand{\bfQ}{\mathbf{Q}}
\newcommand{\bfG}{\mathbf{G}}
\newcommand{\Rp}{{\mathbf{R}^{gc}}}
\newcommand{\Rb}{{\mathbf{R}^{gb}}}
\newcommand{\Rr}{{\mathbf{R}^{cb}}}

\newcommand{\Fg}{\bs{\mathcal{F}}^g}
\newcommand{\Fb}{\bs{\mathcal{F}}^b}
\newcommand{\Fp}{\bs{\mathcal{F}}^c}

% vector symbols
\newcommand{\xg}{\ensuremath{\bs{x}^g}}
\newcommand{\eb}[1]{\ensuremath{\bs{e}^b_{#1}}}
\newcommand{\eg}[1]{\ensuremath{\bs{e}^g_{#1}}}
\newcommand{\et}[1]{\ensuremath{\bs{e}^t_{#1}}}
\newcommand{\ef}[1]{\ensuremath{\bs{e}^f_{#1}}}
\newcommand{\er}[1]{\ensuremath{\bs{e}^r_{#1}}}
\newcommand{\emot}[1]{\ensuremath{\bs{e}^m_{#1}}}
\newcommand{\esc}{\ensuremath{\bs{e}^c_s}}
\newcommand{\eyc}{\ensuremath{\bs{e}^c_y}}
\newcommand{\enc}{\ensuremath{\bs{e}^c_n}}

\newcommand{\xp}{\ensuremath{\bs{x}^p}}
\newcommand{\xps}{\ensuremath{\bs{x}^p_s}}
\newcommand{\xpsmag}{\norm{\xps}}
\newcommand{\eps}{\ensuremath{\bs{e}^p_s}}
\newcommand{\xpy}{\ensuremath{\bs{x}^p_y}}
\newcommand{\xpymag}{\norm{\xpy}}
\newcommand{\epp}{\ensuremath{\bs{e}^p_\perp}}
\newcommand{\xpn}{\ensuremath{\bs{e}^p_n}}
\newcommand{\epn}{\ensuremath{\bs{e}^p_n}}
\newcommand{\xpss}{\ensuremath{\bs{x}^p_{ss}}}
\newcommand{\xpsy}{\ensuremath{\bs{x}^p_{sy}}}
\newcommand{\xpys}{\ensuremath{\bs{x}^p_{ys}}}
\newcommand{\xpyy}{\ensuremath{\bs{x}^p_{yy}}}

\newcommand{\rcom}{\bs{r}_{\text{com}}}
\newcommand{\vcom}{\bs{v}_{\text{com}}}

% scalar symbols
\newcommand{\ths}{\theta^s}
\newcommand{\thp}{\theta^p}
\newcommand{\ks}{\kappa_s^p}
\newcommand{\ky}{\kappa_y^p}

\newcommand{\vb}[1]{\ensuremath{v^b_{#1}}}
\newcommand{\vt}[1]{\ensuremath{v^t_{#1}}}
\newcommand{\vf}[1]{\ensuremath{v^f_{#1}}}
\newcommand{\wb}[1]{\ensuremath{\omega^b_{#1}}}
\newcommand{\wm}[1]{\ensuremath{\omega^m_{#1}}}

\title{Predictive Braking on a Nonplanar Road\thanks{Supported by Brembo Inspiration Lab}}

\author{Thomas Fork\inst{1}\and
Francesco Camozzi\inst{2} \and
Xiao-Yu Fu\inst{2} \and
Francesco Borrelli\inst{1}}

\institute{University of California at Berkeley, Berkeley CA 94708, USA 
\email{\{fork,fborrelli\}@berkeley.edu}
\and
Brembo Inspiration Lab, Sunnyvale CA 94089, USA
\email{\{FCamozzi,XFu\}@us.brembo.com}
}
\maketitle              % typeset the header of the contribution
\begin{abstract}
We present an approach for predictive braking of a four-wheeled vehicle on a nonplanar road. Our main contribution is a methodology to consider friction and road contact safety on general smooth road geometry. We use this to develop an active safety system to preemptively reduce vehicle speed for upcoming road geometry, such as off-camber turns. Our system may be used for human-driven or autonomous vehicles and we demonstrate it with a simulated ADAS scenario. We show that loss of control due to driver error on nonplanar roads can be mitigated by our approach.

\end{abstract}
\section{Introduction}

Nonplanar road geometry plays a major role in the behaviour and safety of ground vehicles that operate in such environments. Operating limits due to road adherence change while new effects appear, such as losing contact when cresting a hill. This paper develops an approach to consider these effects generally on a smooth nonplanar road surface. We develop a novel predictive safety system algorithm for safe vehicle operation on nonplanar road geometry. We show that our safety system maintains safe vehicle speed on a simulated off-camber turn.

This manuscript addresses gaps in current ADAS systems in the treatment of road geometry. Namely, most solutions are designed for flat roads \cite{weigel2006accurate,zhang2018toward}. Approaches that consider more complicated geometry limit their considerations to road curvature, slope, and bank \cite{barreno2022novel,yan2013considering}. These variables are not sufficient to describe roads with curved cross-section and subsequent analysis of vehicle safety is simplified in existing literature. 
The work in \cite{barreno2022novel} ignores changes in vehicle orientation due to road slope and bank when assessing rollover and friction limits. Furthermore, centripetal effects, such as a vehicle driving over a crest or off-camber turn, are absent. The authors of \cite{yan2013considering} consider changes in the components of gravity on a vehicle due to slope but not bank angle, and weight distribution of the vehicle is not considered for rollover prevention. 

This paper presents a new active safety system for predictive braking on nonplanar roads which addresses these shortcomings in a systematic and general manner applicable to general, smooth nonplanar road surfaces.

\section{Vehicle Operating Limits}
\begin{wrapfigure}{R}{0.45\textwidth}
\centering
\vspace{-3em}
\includegraphics[width=0.95\linewidth]{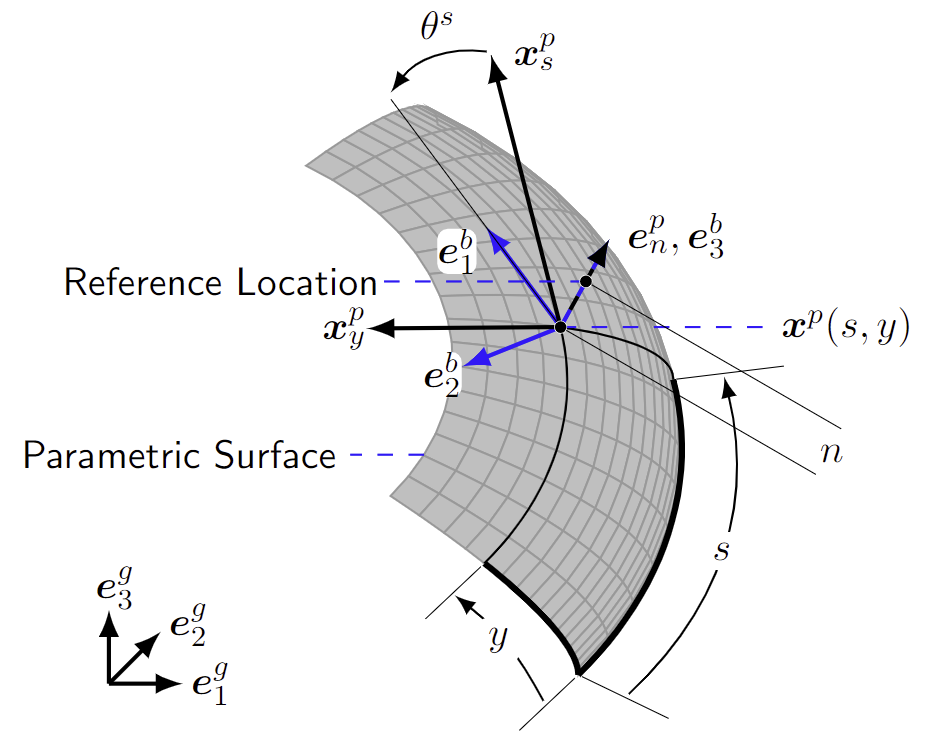}
\caption{Parametric road surface $\xp$. Coordinates $s$, $y$, and $\ths$ describe vehicle pose relative to the surface.}
\label{fig:surf_defn}
\vspace{-2em}
\end{wrapfigure}
We consider three operating limits for a single-body vehicle on a smooth road surface: Tire friction, road contact, and velocity continuity. The last refers to the inability instantly change vehicle speed, and is necessary to anticipate vehicle behaviour on variable road geometry. To consider road geometry in a general sense we use the road model developed in \cite{fork2023models}, which we introduce and use next.

\subsection{Nonplanar Road Model}

The paper \cite{fork2023models} extends the approach of modeling a car as a body tangent to and in contact with a surface to a general parametric surface. The road surface $\xp$ is parameterized by coordinates $s$ and $y$, which then describe a vehicle positioned at normal offset $n$ from the road. Vehicle orientation is described by the angle $\ths$ between the longitudinal vehicle axis $\eb{1}$ and the $s$ tangent vector of the surface: $\xps$. Surface coordinates may be chosen flexibly, such as to follow the center of a lane. The main results hold for any surface parameterization and are:
\begin{subequations} \label{eq:pose_kinematics}
\begin{align}
        \begin{bmatrix}
            \dot{s}\\
            \dot{y}
        \end{bmatrix}
        &=
        \left(\one - n \two\right)^{-1}
        \bfJ
        \begin{bmatrix}
            \vb{1} \\
            \vb{2}
        \end{bmatrix}
    &\label{eq:angular_vel_constraint}
    \begin{bmatrix}
        -\wb{2} \\ \wb{1}
    \end{bmatrix}
    &=\bfJ^{-1} \two\ \paren{\one - n \two}^{-1} \bfJ
    \begin{bmatrix}
    \vb{1} \\
    \vb{2}
    \end{bmatrix}.
\end{align}
\begin{equation}
    \label{eq:ths_dot}
    \dot{\theta}^s =
      \wb{3} +
         \frac{\left(\xpss\times \xps\right)\cdot \xpn}{\xps \cdot \xps}\dot{s}
       + \frac{\left(\xpyy\times \xps\right)\cdot \xpn}{\xps \cdot \xps}\dot{y}
\end{equation}
\end{subequations}
Here $v^b_i$ and $\omega^b_i$ are the ISO body frame components of a vehicle's linear and angular velocity. $\one$ and $\two$ are the first and second fundamental forms of $\xp$, with partial derivatives of $\xp$ denoted by subscripts. $\bfJ$ is the Jacobian between the body frame and $\xp$, used here in the form of a Q-R decomposition:
\begin{subequations}
\begin{align}
    \thp =& - \sin^{-1} \left( \frac{\xps \cdot \xpy}{\xpsmag~\xpymag} \right) & \bfQ &= \begin{bmatrix}
        \xpsmag & 0 \\
        -\sin(\thp) \xpymag & \cos(\thp)\xpymag
    \end{bmatrix}
\end{align}
\begin{equation}
    \bfJ=\begin{bmatrix}
    \xps \cdot \eb{1} & \xps \cdot \eb{2}\\
    \xpy \cdot \eb{1} & \xpy \cdot \eb{2}
    \end{bmatrix}
    =
    \bfQ
    \begin{bmatrix}
        \cos\ths & -\sin\ths\\
        \sin\ths & \cos\ths
    \end{bmatrix}.
\end{equation}
\end{subequations}
The Q-R form for $\bfJ$ simplifies expressions in this manuscript, while Eq. \eqref{eq:pose_kinematics} captures nonplanar behaviour. Coriolis forces and moments on the vehicle will follow from part of the Newton Euler equations:
\begin{align} \label{eq:ne_force}
    F^b_1 &= m\paren{\dot{v}^b_1 - \wb{3}\vb{2}} & F^b_2 &= m\paren{\dot{v}^b_2 + \wb{3}\vb{1}} & F^b_3 &= m \paren{\wb{1}\vb{2}-\wb{2}\vb{1}}
\end{align}
\vspace{-2em}
\begin{align} \label{eq:ne_moment}
    K^b_1 &= I^{b}_{1} \dot{\omega}^b_1 + \left(I^{b}_{3} - I^{b}_{2}\right)\omega^b_2 \omega^b_3 & K^b_2 &= I^{b}_{2} \dot{\omega}^b_2 + \left(I^{b}_{1} - I^{b}_{3}\right)\omega^b_3 \omega^b_1
\end{align}
$\vb{3} = 0$ per the road model \cite{fork2023models} and thus is not present. For motion planning purposes, we will describe vehicle velocity using signed speed $v$, sideslip angle $\beta$, and rates of change of $\ths$ and $\beta$ proportional to $v$ as follows:
\begin{align} \label{eq:vel_assumptions}
    \vb{1} &= v \cos(\beta) & \vb{2} &=v\sin{\beta} & \dot{\theta}^s &= \kappa^s v & \dot{\beta} &= \kappa^\beta v
\end{align}
Expressions for $\dot{v}^b_1$ and $\dot{v}^b_2$ follow via standard calculus. $v^2$ and $\dot{v}$ will be decision variables in our safety system, with $s$, $y$, $\ths$, $\beta$, $\kappa^s$, and $\kappa^\beta$ treated as parameters. These choices will allow our safety system to be implemented as a convex optimization problem. Another result is an expression for $\wb{3}$ from $\dot{\theta}^s$ using \eqref{eq:pose_kinematics}:
\begin{equation} \label{eq:w3_constraint}
    \wb{3} = \kappa^s v - \frac{\left(\xpss\times \xps\right)\cdot \xpn}{\xps \cdot \xps}\dot{s}
       - \frac{\left(\xpyy\times \xps\right)\cdot \xpn}{\xps \cdot \xps}\dot{y}
\end{equation}

\subsection{Friction Cone Constraint}
Using \eqref{eq:ne_force}, constraint \eqref{eq:angular_vel_constraint}, and \eqref{eq:vel_assumptions}, the net vehicle normal force is: 
\begin{equation} \label{eq:net_body_force}
    F^b_3 = mv^2 \begin{bmatrix}\cos(\beta+\ths)&\sin(\beta+\ths)\end{bmatrix} \bfQ^{-1}\two\paren{\one-n\two}^{-1} \bfQ\begin{bmatrix}\cos(\beta+\ths)\\\sin(\beta+\ths)\end{bmatrix}
\end{equation}
$F^b_3$ is linear in $v^2$, meaning the net normal tire force ($F^t_3$) is affine in $v^2$ for a given $s$, $y$, and $\ths$ as gravity forces are constant (found in \cite{fork2023models}) and aerodynamic forces are often approximated as linear in $v^2$. 

Linear expressions for $F^b_1$ and $F^b_2$ follow from the same equation blocks used to derive \eqref{eq:net_body_force} and are not expanded here. As a result, net longitudinal and lateral tire forces $F^t_1$ and $F^t_2$ are affine in $\dot{v}$ and $v^2$ by the same assumptions. The complete friction cone constraint is then:
\begin{equation} \label{eq:friction_cone_constraint}
    \norm{\begin{matrix}F^t_1 & F^t_2 \end{matrix}}_2 \leq \mu F^t_3
\end{equation}
where $\mu$ is a road adherence parameter. This constraint is a convex second order cone constraint as the tire forces are affine functions of $\dot{v}$ and $v^2$.

\subsection{Road Contact Constraint} \label{sec:road_contact}

\begin{wrapfigure}{r}{0.45\textwidth}
\vspace{-4em}
\centering
\includegraphics[width=0.95\linewidth]{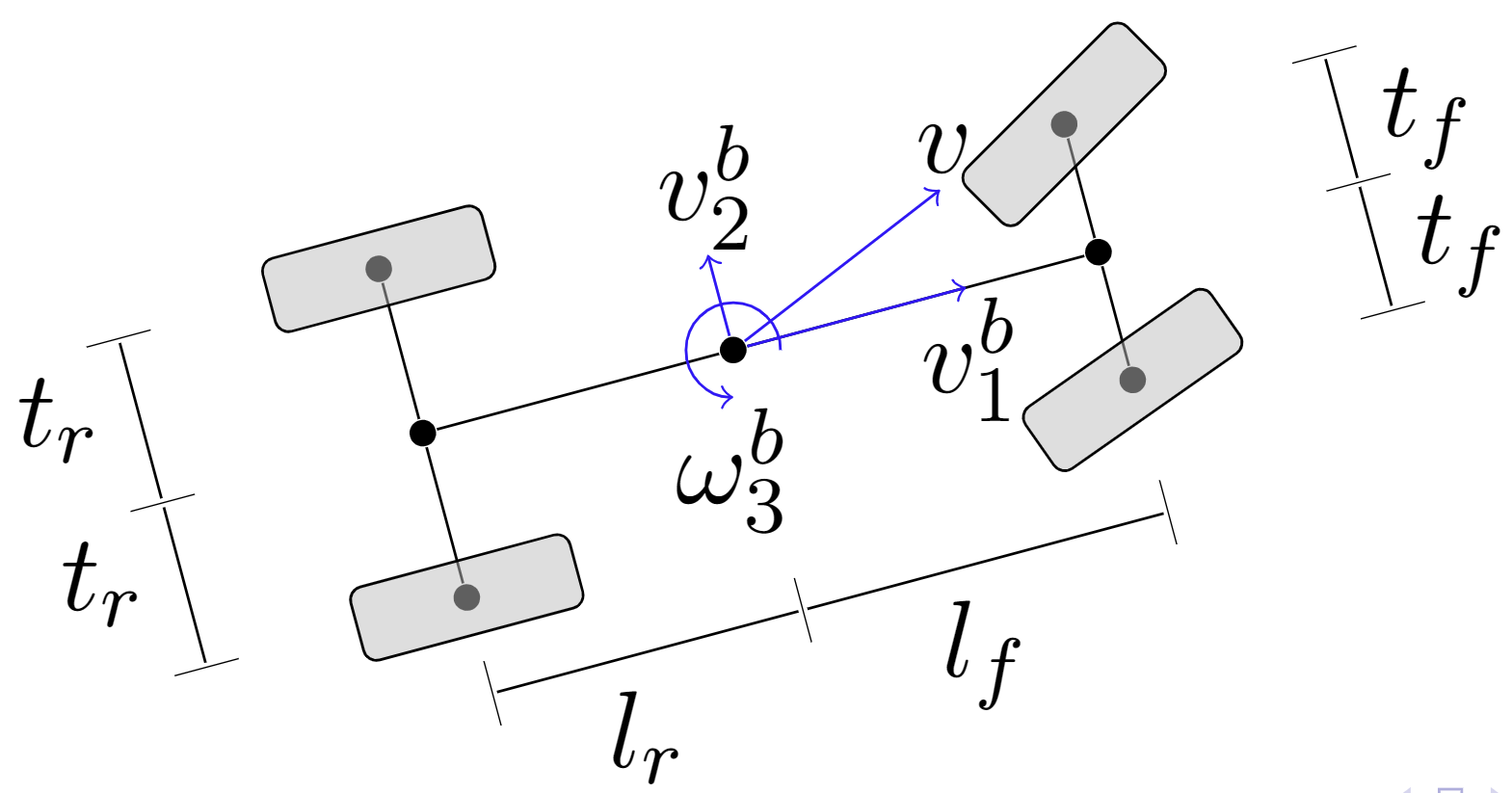}
\caption{Wheelbase dimensions used for weight distribution.}
\label{fig:wheelbase}
\end{wrapfigure}

Enforcing road contact requires modeling weight distribution, which requires modeling the roll and pitch moment on the vehicle. These follow from \eqref{eq:ne_moment}, where the $\wb{1,2,3}$ coefficients are linear in $v$ due to \eqref{eq:angular_vel_constraint} and \eqref{eq:w3_constraint}. For $\dot{\omega}^b_{1,2}$ we use the approximation from \cite{fork2023models} that 
\begin{equation} \label{eq:angular_accel_constraint}
    \begin{bmatrix}
        -\dot{\omega}_2^b \\ \dot{\omega}_1^b
    \end{bmatrix}
    \approx \bfJ^{-1} \two~\left(\one - n \two\right)^{-1} \bfJ
    \begin{bmatrix}
    \dot{v}_1^b \\
    \dot{v}_2^b
    \end{bmatrix}.
\end{equation}

Expansion of \eqref{eq:ne_moment} using \eqref{eq:pose_kinematics}, \eqref{eq:vel_assumptions}, \eqref{eq:w3_constraint}, and \eqref{eq:angular_accel_constraint} provides expressions for roll and pitch moments $K^b_1$ and $K^b_2$. These are linear in $v^2$ and $\dot{v}$ and omitted for brevity. For weight distribution we consider moments from tire normal forces. The dominant source of other moments are longitudinal and lateral tire forces, which produce moments about the height of the center of mass $h$. Moments due to tire normal forces $K^N_1$ and $K^N_2$ are then:
\begin{align}
    K^N_1 &= K^b_1 - F^t_2 h & K^N_2 &= K^b_2 + F^t_1 h
\end{align}
These are affine in $v^2$ and $\dot{v}$, and may be extended to include $v^2$ terms for aerodynamic moments. To model the forces on individual tires, we use the load-transfer model of \cite{rucco2014development} with wheelbase dimensions in Figure \ref{fig:wheelbase}:
\begin{subequations}\label{eq:weight_distribution}
\begin{align} 
    N_f &= \frac{F^t_3 l_r - K^N_2}{l_r+l_f} & 
    N_r &= \frac{F^t_3 l_f + K^N_2}{l_r+l_f} & 
    \delta &= \frac{K^N_1}{2\paren{t_f^2 + t_r^2}}
\end{align}
\vspace{-1em}
\begin{equation}
    \begin{aligned}
    N_{fr} &= N_f - \delta t_f &
    N_{fl} &= N_f + \delta t_f &
    N_{rr} &= N_r - \delta t_r &
    N_{rl} &= N_r + \delta t_r
    \end{aligned}
\end{equation}
\end{subequations}
The four tire normal forces $N_{fr}$ (front right) through $N_{rl}$ (rear left) are affine in $v^2$ and $\dot{v}$, meaning that constraining them to be positive to avoid loss of road contact is a convex constraint:
\begin{align} \label{eq:normal_force_constraint}
    N_{fr} & \geq 0 & N_{fl} & \geq 0 & N_{rr} & \geq 0 & N_{rl} & \geq 0
\end{align}

\subsection{Velocity Continuity Constraints}
To develop our safety system, friction cone and road contact constraints are introduced at fixed points in space in a multistage control problem presented next. These stages must be connected together with velocity constraints relating $v^2$ and $\dot{v}$ at adjacent stages to capture vehicle speed changing over time. We use a midpoint integration scheme similar to \cite{verscheure2009time}:
\begin{equation} \label{eq:vel_continuity}
    \paren{v^2}^{k+1} = \paren{v^2}^k + \frac{1}{2}\paren{\dot{v}^k + \dot{v}^{k+1}} \paren{l^{k+1} - l^{k}}
\end{equation}
Here superscript $^k$ denotes the stage of the control problem, and $l^k$ is the arc length traveled by a vehicle to reach stage $k$.

\section{Active Safety System}
\begin{wrapfigure}{r}{0.55\textwidth}
\vspace{-4em}
\begin{subequations}\label{eq:ssop}
\begin{align} 
    \min_{\paren{v^2}^k,\dot{v}^k} & \sum_{k=0}^{N-1} \abs{\paren{F^t_1}^k - B} \label{eq:obj_function}\\
    \text{subject to  } & \text{Eq. \eqref{eq:friction_cone_constraint}} & \forall \ k\\
    & \text{Eq. \eqref{eq:normal_force_constraint}} & \forall \ k \\
    &\text{Eq. \eqref{eq:vel_continuity}} & \forall \ k \\
    & \paren{v^2}^0 = \paren{v_0}^2 \label{eq:initial_speed_constraint}
\end{align}
\end{subequations}
\vspace{-2em}
\end{wrapfigure}

We develop an active safety system with safety limits encoded by constraints \eqref{eq:friction_cone_constraint}, \eqref{eq:normal_force_constraint}, and \eqref{eq:vel_continuity}.
These constraints are convex in $v^2$ and $\dot{v}$ for fixed $s$, $y$, $\ths$, $\beta$, $\kappa^s$, and $\kappa^\beta$, which we respectively make decision variables and parameters for optimization problem \eqref{eq:ssop}. 
We introduce stages $0$ through $N-1$ with decision variables and parameters present at each stage. \eqref{eq:vel_continuity} constrains the speed between adjacent stages, while \eqref{eq:friction_cone_constraint} and \eqref{eq:normal_force_constraint} constrain individual stages. We introduce parameter $v_0$ for initial speed of the vehicle, and $B$ for a nominal brake force input. We use the objective function \eqref{eq:obj_function} which is the total absolute difference between $B$ and the longitudinal tire force at each stage. 

The main output of this optimization problem is $F^t_1 - B$ at each stage, which informs how much tire forces must change relative to $B$ for continued safe vehicle operation. As an example application, $B$ could be a pedal request from a driver and $F^t_1 - B$ being nonzero indicates active safety measures must be taken, such as an automated brake procedure. Minimizing $F^t_1 - B$ corresponds to intervening only when necessary, such as if a driver fails to slow down for an off-camber turn. 

We note that the core novelty of this safety system is the nonplanar road safety constraints. These are not limited to speed-limiting applications, and may be used for active steering, suspension, and powertrain systems as well.

\section{Simulation Environment}
We tested our active safety system using a simulated lane-keeping scenario on a nonplanar road surface. We used a nonplanar two-track vehicle model with suspension motion based on \cite{fork2023models} with a combined-slip Pacejka tire model \cite{pacejka_tire_book}. Driver behaviour was simulated with a PI steering controller. Brake actuators and slip control were simulated with a proprietary Brembo model. We implemented our safety system \eqref{eq:ssop} with parameters for each stage corresponding to following the center of a lane, with brake force targets handled by a nonplanar electronic brakeforce distribution (EBD) algorithm described here:

The core component of EBD is distributing a target brake force and moment over the four wheels of a car. This is fundamentally limited by the road adherence of each tire and any limits of the brake actuators. We consider weight distribution effects in a general manner by careful use of accelerometer data. The raw output of any accelerometer is proper acceleration, which is related to coordinate acceleration in an inertial frame via gravitational acceleration $\bs{g}$.
\begin{equation}
    \bs{a}_{\text{proper}} = \bs{a}_{\text{coordinate}} - \bs{g} = \frac{1}{m} \bs{F}^b - \bs{g}
\end{equation}
The far-right expression follows from Newtonian mechanics, telling us the accelerometer measures every force but gravity. We use this to estimate tire force components directly, which are used to compute the net normal force and moments in \eqref{eq:weight_distribution}. The four normal forces then inform our EBD algorithm, which distributes target net brake force and moment over the four wheels. 

Two test cases were considered: First, impulsive brake application after an initial delay, modeling a delayed driver. Second, driver brake application was removed and the test repeated with the active safety system present. All simulations used the same steering control, brake control, vehicle simulator, and road surface: a u-turn with a 30\% off-camber bank.

\begin{figure}
    \centering
    \begin{subfigure}{0.48\linewidth}
        \centering
        \includegraphics[width=\linewidth]{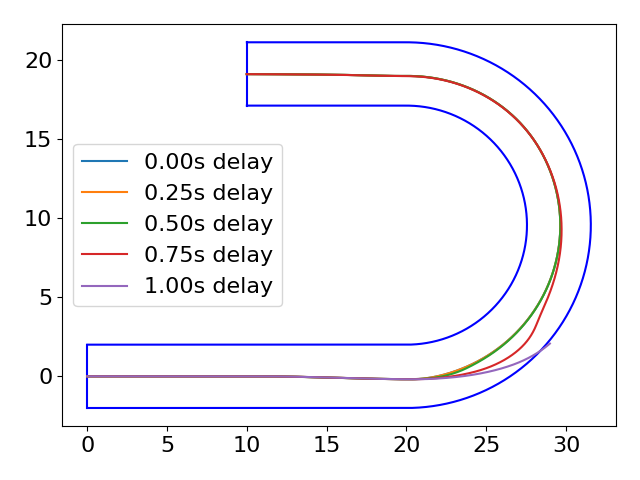}
        \caption{No safety system and delayed driver}
        \label{fig:results_ol}
    \end{subfigure}%
    \hfill
    \begin{subfigure}{0.48\linewidth}
        \centering
        \includegraphics[width=\linewidth]{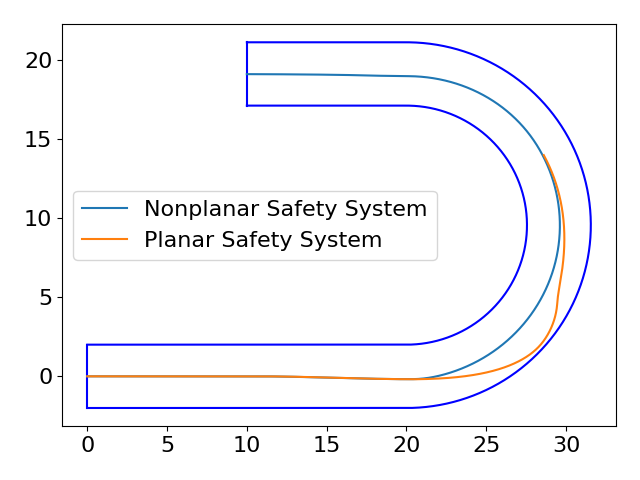}
        \caption{Safety system active}
        \label{fig:results_ss}
    \end{subfigure}%
    \caption{Vehicle trajectories on nonplanar u-turn, starting from bottom left}
    \label{fig:results}
\end{figure}

\section{Results and Conclusion}
Closed loop vehicle trajectories with and without safety system are shown in Figure \ref{fig:results}. As evidenced in Figure \ref{fig:results_ol}, a proactive driver can maintain control of the vehicle, but must brake almost immediately to follow the lane. With the nonplanar safety system (Figure \ref{fig:results_ss}) no longitudinal driver intervention is necessary. However, implementation of the same system with a planar road model results in loss of control. Our safety system mitigates loss of control of a vehicle using knowledge of the road surface and intended vehicle motion.

\bibliographystyle{splncs04}
\bibliography{main}

\begin{thebibliography}{1}
\providecommand{\url}[1]{\texttt{#1}}
\providecommand{\urlprefix}{URL }
\providecommand{\doi}[1]{https://doi.org/#1}

\bibitem{barreno2022novel}
Barreno, F., Santos, M., Romana, M.G.: A novel adaptive vehicle speed recommender fuzzy system for autonomous vehicles on conventional two-lane roads. Expert Systems p. e13046 (2022)

\bibitem{fork2023models}
Fork, T., Tseng, H.E., Borrelli, F.: Models for ground vehicle control on nonplanar surfaces. Vehicle System Dynamics pp. 1--25 (2023)

\bibitem{pacejka_tire_book}
Pacejka, H.: Tire and Vehicle Dynamics. Butterworth-Heinemann, 3rd edn. (2012)

\bibitem{rucco2014development}
Rucco, A., Notarstefano, G., Hauser, J.: Development and numerical validation of a reduced-order two-track car model. European Journal of Control  \textbf{20}(4),  163--171 (2014)

\bibitem{verscheure2009time}
Verscheure, D., Demeulenaere, B., Swevers, J., De~Schutter, J., Diehl, M.: Time-optimal path tracking for robots: A convex optimization approach. IEEE Transactions on Automatic Control  \textbf{54}(10),  2318--2327 (2009)

\bibitem{weigel2006accurate}
Weigel, H., Cramer, H., Wanielik, G., Polychronopoulos, A., Saroldi, A.: Accurate road geometry estimation for a safe speed application. In: 2006 ieee intelligent vehicles symposium. pp. 516--521. IEEE (2006)

\bibitem{yan2013considering}
Yan, X., Zhang, R., Ma, J., Ma, Y., et~al.: Considering variable road geometry in adaptive vehicle speed control. Mathematical Problems in Engineering  \textbf{2013} (2013)

\bibitem{zhang2018toward}
Zhang, Y., Chen, H., Waslander, S.L., Yang, T., Zhang, S., Xiong, G., Liu, K.: Toward a more complete, flexible, and safer speed planning for autonomous driving via convex optimization. Sensors  \textbf{18}(7), ~2185 (2018)

\end{thebibliography}
\end{document}